\documentclass[conference]{IEEEtran}

\usepackage{epsfig}
\usepackage{epstopdf}
\usepackage{graphicx}
\usepackage{amsmath}
\usepackage{amssymb}
\usepackage{array}
\usepackage{algorithm}
\usepackage{algorithmic}
\usepackage{hhline}
\newcolumntype{C}[1]{>{\centering\arraybackslash}p{#1}}

\ifCLASSINFOpdf
\else
\fi
\begin{document}
%

\title{Automatic Building Extraction in Aerial Scenes Using Convolutional Networks}

\author{
\IEEEauthorblockN{Jiangye Yuan}
       \IEEEauthorblockA{Computational Sciences \& Engineering Division\\
       Oak Ridge National Laboratory\\
       Oak Ridge, Tennessee 37831\\
       Email: yuanj@ornl.gov}
}

\maketitle

\begin{abstract} 
Automatic building extraction from aerial and satellite imagery is highly challenging due to extremely large variations of building appearances. To attack this problem, we design a convolutional network with a final stage that integrates activations from multiple preceding stages for pixel-wise prediction, and introduce the signed distance function of building boundaries as the output representation, which has an enhanced representation power. We leverage abundant building footprint data available from geographic information systems (GIS) to compile training data. The trained network achieves superior performance on datasets that are significantly larger and more complex than those used in prior work, demonstrating that the proposed method provides a promising and scalable solution for automating this labor-intensive task.
\end{abstract} 

\section{Introduction}
Up-to-date maps of buildings are crucial for navigation, urban planning, disaster management, population estimation, and many other geospatial related applications. As advances are made in remote sensing technologies, high resolution overhead imagery including spaceborne and airborne images is widely available and thus provides an ideal data source for creating such maps. However, manually delineating buildings on images is notoriously time and effort consuming. 

Due to the potential productivity gain, automatic building extraction has been extensively studied for decades. Much of the past work defines criteria of building appearance such as uniform colors, regular shapes, and nearby shadows, and designs a system that identifies objects satisfying the criteria \cite{kim1999,inglada2007,cote2013,li2015}. Such approaches have limited generalization abilities because the criteria account for only certain types of buildings whether hand-coded or learned from samples, and results are sensitive to parameter choice. Therefore, despite promising performance on small images containing relative homogeneous buildings, at the time of writing there has not been any automatic system that works reliably on real-world datasets (e.g., images containing a large urban scene). Some methods utilize LiDAR data that give detailed height information and are able to obtain more reliable results \cite{sohn2007,zhou2008,awrangjeb2010}. However, compared with imagery, LiDAR data are considerably more expensive to acquire and thus much less accessible. In this paper, we focus on image data.  

Recognizing the difficulty in developing automatic methods, crowdsourcing emerges as an alternative strategy. One of the most successful examples is OpenStreetMap\footnote{http://www.openstreetmap.org}, which has millions of contributors providing manual labeling. However, a major problem of crowdsourcing maps is inconsistent quality. Position accuracy and completeness vary greatly across different places due to participation inequality \cite{nielsen2006}. The issue is more severe for buildings than other objects like roads and water bodies, because buildings are smaller objects and require more effort to identify and delineate.     

We take a unique approach that utilizes the convolutional network (ConvNet) framework coupled with labeled data procured from GIS resources. Deep ConvNets trained with very large labeled data have shown to be very powerful to capture the hierarchical nature of features in images and generalize beyond training samples \cite{lecun2015}. The capabilities lead to great success in many challenging pattern recognition tasks \cite{krizhevsky2012,tompson2014,sainath2013}. Meanwhile, there exist a massive amount of building footprints from various GIS databases, including crowdsourcing resources. Since both building footprints and images are georeferenced, they can easily be converted to training samples where individual buildings are labeled. Unlike many machine learning problems that requires tremendous effort to collect enough labeled data, abundant human-labeled data are readily available for building extraction. 

The remainder of the paper is organized as follows. In Section~\ref{sec:netarch}, we present a network architecture capable of learning pixel-wise classification. Section~\ref{sec:outrep} introduces an output representation that is well suited for our task. In Section~\ref{sec:expm}, we provide the details of data preparation and network training, as well as evaluation on large datasets. Finally, we conclude in Section~\ref{sec:ccl}. 

\section{Network architecture}
\label{sec:netarch}

A typical ConvNet has a series of stages. In each stage, a set of feature maps are convolved with a bank of filters that are subject to training, and filter responses are passed through some non-linear activation function to form new feature maps, which are downsampled via a pooling unit to generate output maps with reduced spatial resolution. As traveling through stages, an input image is transferred into coarser representations that encode higher-level features. In the final stage, the output feature maps constitute highly discriminative representation for whole-image classification.  

Mapping buildings from images is essentially a problem of segmenting semantic objects. While output feature maps from a ConvNet capture high-level semantic information, such representations discard spatial information at fine resolution that is crucial for dense prediction. In contrast to image classification problems where ConvNet approaches are well established, how to effectively apply ConvNets to segmentation is still being explored. Recent work suggests a number of methods. Farabet et al. \cite{farabet2013} assign patch-wise predictions from a ConvNet to superpixels and merge superpixels into meaningful regions through a Conditional Random Field (CRF) method. Multiple networks with shared parameters are exploited in \cite{pinheiro2014}, where dense predictions are achieved by shifting input images and merging outputs. Silberman et al. \cite{silberman2014} find regions corresponding to individual objects from a segmentation hierarchy, where the feature of a region is computed by feeding its bounding box to a pre-trained ConvNet. In \cite{long2015}, the coarse predictions from a late stage are upsampled and combined with predictions from its preceding stage to generate denser predictions, which are again upsampled and combined with predictions in an earlier stage. The procedure continues until reaching a desired resolution.   

Although these methods show promise in segmenting natural images, they have components not suited for building extraction. First, certain segmentation methods are employed to provide initial partitions or refine results \cite{farabet2013,silberman2014}. It has been recognized that segmentation methods that designed for natural images are not guaranteed to work for remote sensing images, which have different data characteristics as well as much larger image sizes. Second, training on small patches \cite{farabet2013, pinheiro2014} is problematic because in high resolution remote sensing images such patches tend to cover fragmented buildings and thus fail to capture complete information of individual buildings. Third, prediction density is limited. For instance, the output from \cite{long2015} is at one-eighth of the input resolution. A low resolution output sacrifices positioning precision given by high resolution images. Last but definitely not least, most of the methods assign the same label to objects of the same class, where individual objects cannot be separated if they form connected regions, while it is important to identify individual buildings that are adjacent to others. 

Feature maps from different stages capture local information in different neighborhood sizes and at different semantic levels. The first stage generates feature maps with fine spatial resolution, where each unit looks at a small portion of input image (receptive fields) and responds to low level features like edges and corners. The last stage outputs very coarse feature maps, where each unit pools information from a large subwindow and detects objects composed of multiple parts. Feature maps from the other stages correspond to certain intermediate levels in the feature hierarchy. For building extraction, these feature maps need to be used jointly to decide whether a pixel is on buildings or not. For example, a fine resolution feature map capturing edges is necessary for precise boundary localization, and a coarse resolution feature map characterizing large neighborhoods can differentiate between a flat rooftop and an empty lot on the ground. 

\begin{figure}
\begin{center}
\includegraphics[width=0.4\textwidth]{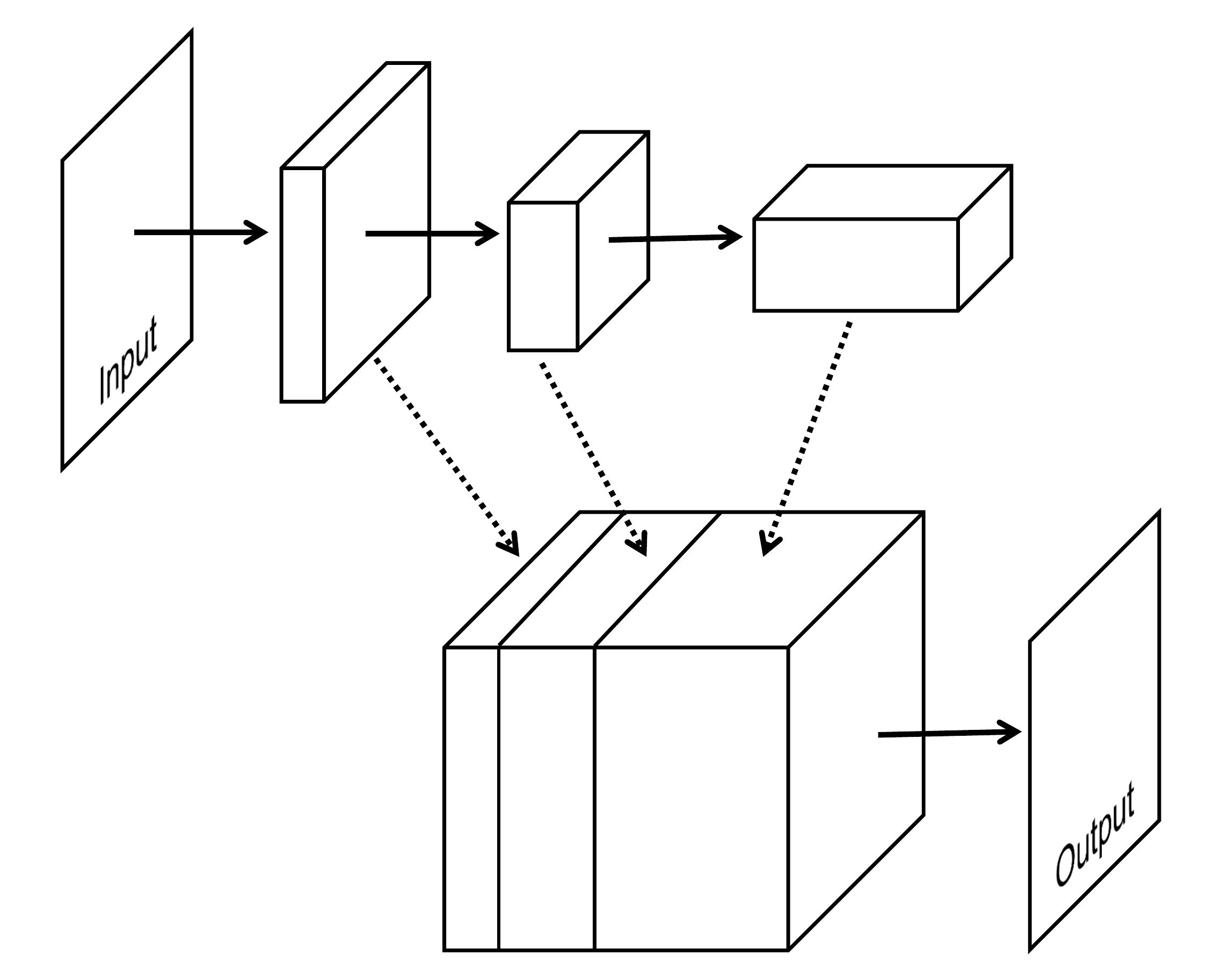}
\end{center}
\caption{A ConvNet integrating multi-stage feature maps. Solid arrows represent convolutional operations, and dotted arrows upsampling. \label{fig:brchnet}}
\end{figure}

To achieve pixel level classification, we propose to combine the outputs from multiple stages through a network architecture illustrated in Fig.~\ref{fig:brchnet}. Here we focus on the structure for combining multi-stage feature maps. Detailed network configurations will be discussed in the experiment section. In the network, each of the first three stages contains a convolutional layer followed by a pooling layer. The output of the last stage is upsampled to the same size of input. The output of the first two stages are branched and upsampled to the input size. Upsampling is achieved via bilinear interpolation. Upsampled features maps are stacked and fed into a convolutional layer with a filter of size $1 \times1 \times n$ ($n$ denotes the number of stacked feature maps) to generate an output map. This convolutional layer is equivalent to a single perceptron layer applied to activations across feature maps corresponding to the same pixel location. Hence, the network outputs pixel-wise labels predicted based on information from multiple stages. 

Compared to regular ConvNets, the main difference of the proposed network is that output of a stage is branched and an upsampling operation of bilinear interpolation is employed. In order to determine whether such a network can be trained by the backpropagation algorithm, we need to analyze how the two differences affect gradient computation. Branching can be addressed by the multivariable chain rule, which states that if a variable branches out to different paths, then partial derivatives from all paths should be added. In the first two stages of the network in Fig.~\ref{fig:brchnet}, output feature maps go to a convolution operation and an unsampling operation. During backpropagation, the partial derivatives of both operations are summed and passed to the preceding operation.  

\begin{figure}
\begin{center}
\includegraphics[width=0.43\textwidth]{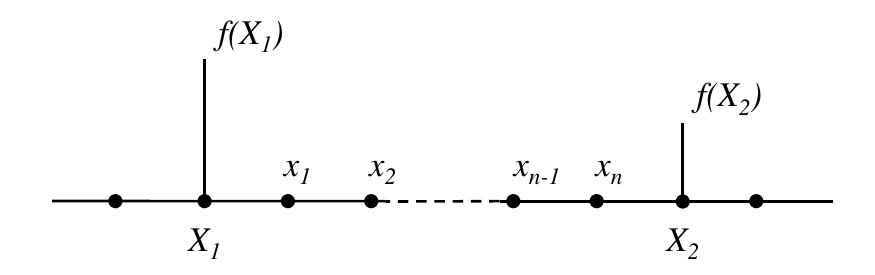}
\end{center}
\caption{An illustration of linear interpolation. \label{fig:bli}}
\end{figure}

It is not straightforward to find out the derivatives of bilinear interpolation. Here we show that bilinear interpolation can be reformulated as convolution, the derivative of which is well understood. Without loss of generality, let us consider one dimensional interpolation illustrated in Fig.~\ref{fig:bli}, where $X_1$ and $X_2$ are two points with known values $f(X_1)$ and $f(X_2)$ in an one-dimensional signal, and the goal is to compute the values at $n$ equally-spaced points in between to obtain an upsampled signal. For simplicity, we omit points in the same setting spreading to both sides. According to the definition of linear interpolation, any of the $n$ points, $x_i$, takes the value of 
\begin{equation} 
f(x_i) = \frac{X_2-x_i}{X_2 - X_1} f(X_1) + \frac{x_i-X_1}{X_2 - X_1} f(X_2),
\label{blidef}
\end{equation} 
which is the sum of two values with inversed distance weighting. 

Now we take the original signal and set the unknown values to zeros. We convolve the signal with a filter in the form of 
\begin{equation}
[\frac{1}{n+1}, \frac{2}{n+1}, \dots, \frac{n+1}{n+1}, \dots, \frac{2}{n+1}, \frac{1}{n+1}]. \label{blifilter}
\end{equation} 
At both $x_1$ and $x_2$, the values after convolution remain unchanged because the filter has one at its center and does not cover any neighboring points with known values. At $x_i$, the convolution gives 
\begin{equation}
f'(x_i) = \frac{n+1-i}{n+1} f(X_1) + \frac{i}{n+1} f(X_2).
\label{bliconv}
\end{equation} 
Let $L$ denotes the length between $x_i$ and $x_{i+1}$. Since $X_2-x_i$ equals to $(n+1-i)L$, $x_i-X_1$ equals to $iL$, and $X_2-X_1$ equals to $(n+1)L$, we obtain $f(x_i) = f'(x_i)$. Accordingly, bilinear interpolation can be achieved via convolution with a special filter,\footnote{Two-dimensional interpolation can conveniently be handled by applying the same filter in both directions.} the derivatives of which are therefore convolution with the same filter. Worthing mentioning is that the reformulation also provides a simple and efficient implementation of  bilinear interpolation using convolution, a well optimized operation in many computational libraries.


We notice that Hariharan et al. \cite{hariharan2014} use a similar structure in the last stage of ConvNets for natural image segmentation. However, their network needs to take input of a bounding box enclosing objects, which is generated by some object detector, and outputs a coarse label map of size $50 \times 50$ pixels, while our network works on raw pixels without any prior detection step and produces results at a high resolution. Moreover, in \cite{hariharan2014} there is no discussion on how the training algorithm works with new components in the network architecture.

Our network involves only local operations and hence can be used for arbitrary-sized input. This is a highly desired property for processing remote sensing images. During training we tend to use relatively small image tiles, which increase randomness and are memory friendly. When applying the network to new images, since the border effect caused by filtering makes results near image borders unreliable, special treatments are needed to combine results of small tiles into a large one. The proposed network is capable of taking large images as a whole, which greatly reduces the chance of encountering border effects.

\section{Output representation}
\label{sec:outrep}

\begin{figure}
\begin{center}
\includegraphics[width=0.21\textwidth]{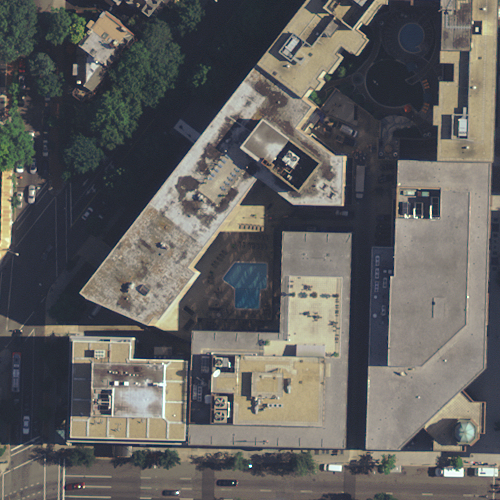}
\hspace{2mm}
\includegraphics[width=0.21\textwidth]{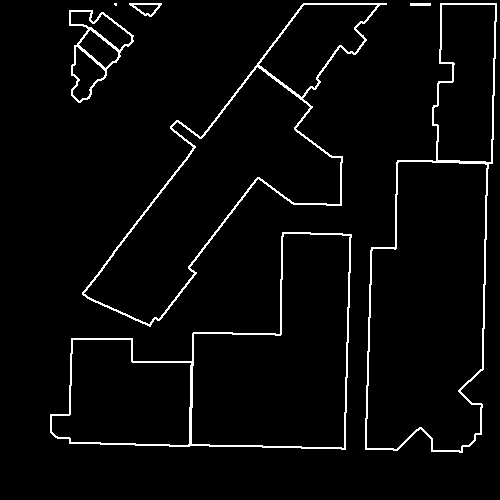}
\makebox[0.21\textwidth][c]{(a)}
\hspace{2mm}
\makebox[0.21\textwidth][c]{(b)}

\includegraphics[width=0.21\textwidth]{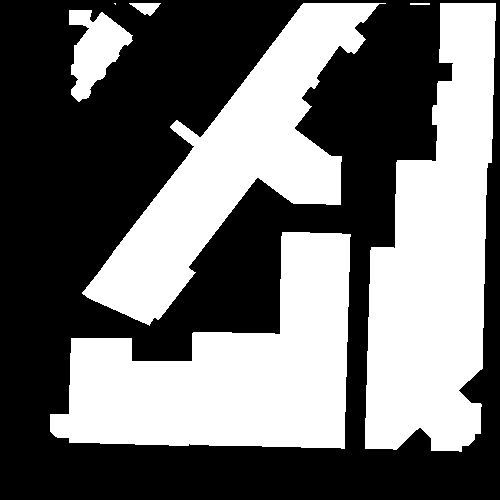}
\hspace{2mm}
\includegraphics[width=0.21\textwidth]{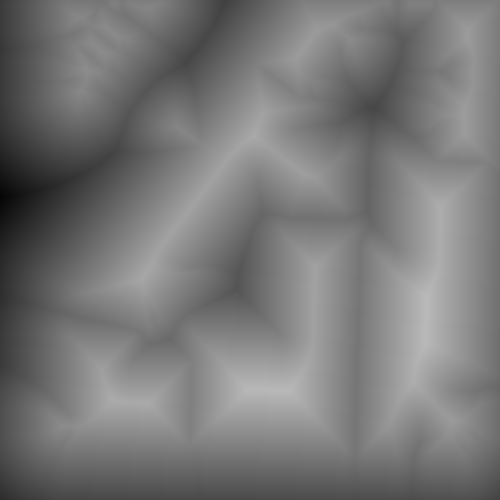}
\makebox[0.21\textwidth][c]{(c)}
\hspace{2mm}
\makebox[0.21\textwidth][c]{(d)}
\end{center}
\caption{Output representation. (a) Original image. (b) Boundary map. (c) Region map. (d) Signed distance transform. Gray values are proportional to distance values. \label{fig:outPres}}
\end{figure}

For segmenting buildings, there are two commonly used output representations, boundary maps that label pixels corresponding to building outlines, and region maps that label building pixels (see Fig.~\ref{fig:outPres}(a)-(c)). A network trained on labeled data in either form is adapted to the corresponding output representation. However, both representations have inherent deficiencies. A network producing boundary maps is a contour detector, which does not guarantee closed contours and hence may produce significantly misshaped building outlines. More importantly, presenting labeled data as boundary maps abandons valuable information of whether pixels are inside or outside buildings. Although in this aspect training on region maps makes better use of labeled data, region maps cannot represent boundaries of adjacent buildings, and thus such boundaries can be neither exploited during training nor detected in test. In experiments, we observe that networks trained on labeled data in either representation are difficult to converge and yield dissatisfying results.  

It is desirable to have a representation that encodes all boundaries as well as region masks. The signed distance function provides a well-suited option. The signed distance function value at a pixel is equal to the distance from the pixel to its closest point on boundaries with positive indicating insider objects and negative otherwise. Fig.~\ref{fig:outPres}(d) illustrates the signed distance representation of building footprints. The advantage of this representation is twofold. First, boundaries and regions are captured in a single representation and can be easily read out. Building outlines are pixels with values equal to zero, and buildings correspond to positive valued pixels. Second, labeled data transformed via the signed distance function can be regarded as fine-grained labels, where pixels are assigned to finely divided classes based on their distance to boundaries instead of a small number of coarse classes (e.g., building and non-building). Such labeled data force a network to learn more informative cues (e.g., the difference between pixels near boundaries and those far away), which can benefit classification performance. 

\section{Experiments}
\label{sec:expm}

\subsection{Data preparation}
In our experiments, we use a collection of aerial images at 0.3 meter resolution covering the entire Washington D.C. area and the corresponding building footprint layer in form of vector data downloaded from a public GIS database\footnote{http://opendata.dc.gov/}. Although the building footprints are produced using data sources different from the images, they are mostly consistent with each other. To train the network, we create image tiles of size $500 \times 500$ pixels. In order to have image tiles with fewer partial buildings, the following procedure is performed. For each building polygon in the vector data, we move a 150 m $\times$ 150 m window (the same as an image tile covers) around the center of the building. The area with the fewest buildings across borders is selected, and all the buildings within the area are overlaid with the corresponding image tile. It is very often that GIS data are not well aligned with images. We follow the method used in \cite{yuan2014}, which is to compute a cross-correlation between building footprints and image gradients. Building footprints are shifted to the place where the correlation coefficient reaches the maximum. The alignment between two data sources can be greatly improved, hence a better quality of labeled data. The training dataset consists of 2000 image tiles and the corresponding building masks.  For testing, we use ten $3000 \times 3000$ image tiles covering areas excluded from the training data. The test data includes around 7,400 buildings, an order of magnitude larger than what were used in previous work \cite{benedek2012,cote2013,li2015}.  

\subsection{Network configuration and training}
Our network contains seven regular ConvNet stages and a final stage for pixel level classification. The first stage applies 50 filters of size $5 \times 5 \times 3$ to a $500 \times 500 \times 3$ input image and performs non-overlapping max-pooling over a $2 \times 2$ unit region. Each of the following three stages has a similar structure containing both a convolutional layer and a max-pooling layer. Convolutional layers have 70 filters of size $5 \times 5 \times 50$, 100 filters of size $3 \times 3 \times 70$, and 150 filters of size $3 \times 3 \times 100$, respectively. Each of the next three stages filters the output from its preceding stage and produces feature maps without max-pooling, where convolutional layers has 100 filters of size of $3 \times 3 \times 150$, 70 filters of size of $3 \times 3 \times 100$, and 70 filters of size of $3 \times 3 \times 70$, respectively. In feature maps output from the seventh stage, each unit has an effective receptive field of $148\times148$, which is sufficiently large to cover most individual buildings in the dataset. See the Appendix for a method to calculate receptive field sizes. The final stage takes output from the first, second, third, and seventh stages for pixel-wise prediction using the structure described in Section \ref{sec:netarch}. The Rectified Linear Unit (ReLU) nonlinearity is used in all convolutional layers.

In the final stage, all feature maps are upsampled to $250 \times 250$, which is the highest resolution of feature maps (from the first stage), and stacked together. The resulting pixel labels have the same resolution, i.e., half the original image resolution, which is practically adequate. We apply 128 filters of size of $1 \times 1 \times 290$ to the feature map stack. As a result, each pixel has a prediction vector similar to class distribution in multi-class classification, which is then normalized by the softmax function. Each element of the normalized vector indicates the probability of the pixel within a certain distance range. In training, we minimize the cross entropy with labeled data that are truncated and quantized to the 128 integers from -64 to 63. In test, the result at each pixel is the sum of the 128 integers weighted by the normalized prediction vector, which can be regarded as the expectation of output variable. An alternative to generate pixel prediction is to apply only one filter to the feature map stack, and each pixel has a single-value prediction that can be used as the final result. However, we observe that such a network easily saturates during training. 

The network is trained in an end-to-end manner. No pre- or post-processing is used. We train the networking using stochastic gradient descent with 5 images as a mini-batch. The weight update rule in \cite{krizhevsky2012} is used with learning rate 0.02, momentum 0.9, and weight decay $5^{-5}$. The weights of all filters are initialized with uniform random numbers, and all biases are initialized with zeros. 200 randomly selected images from the training dataset are used for validation with the metric of average misclassification rate.    

The implementation is based on the Theano library \cite{Bastien2012}. We train the network on a single NVIDIA Tesla 6GB GPU. The training was stopped after 800 epochs, which took roughly two days. The code will be made available to the public. 

\subsection{Results}
\begin{figure*}
\begin{center}
\includegraphics[width=0.48\textwidth]{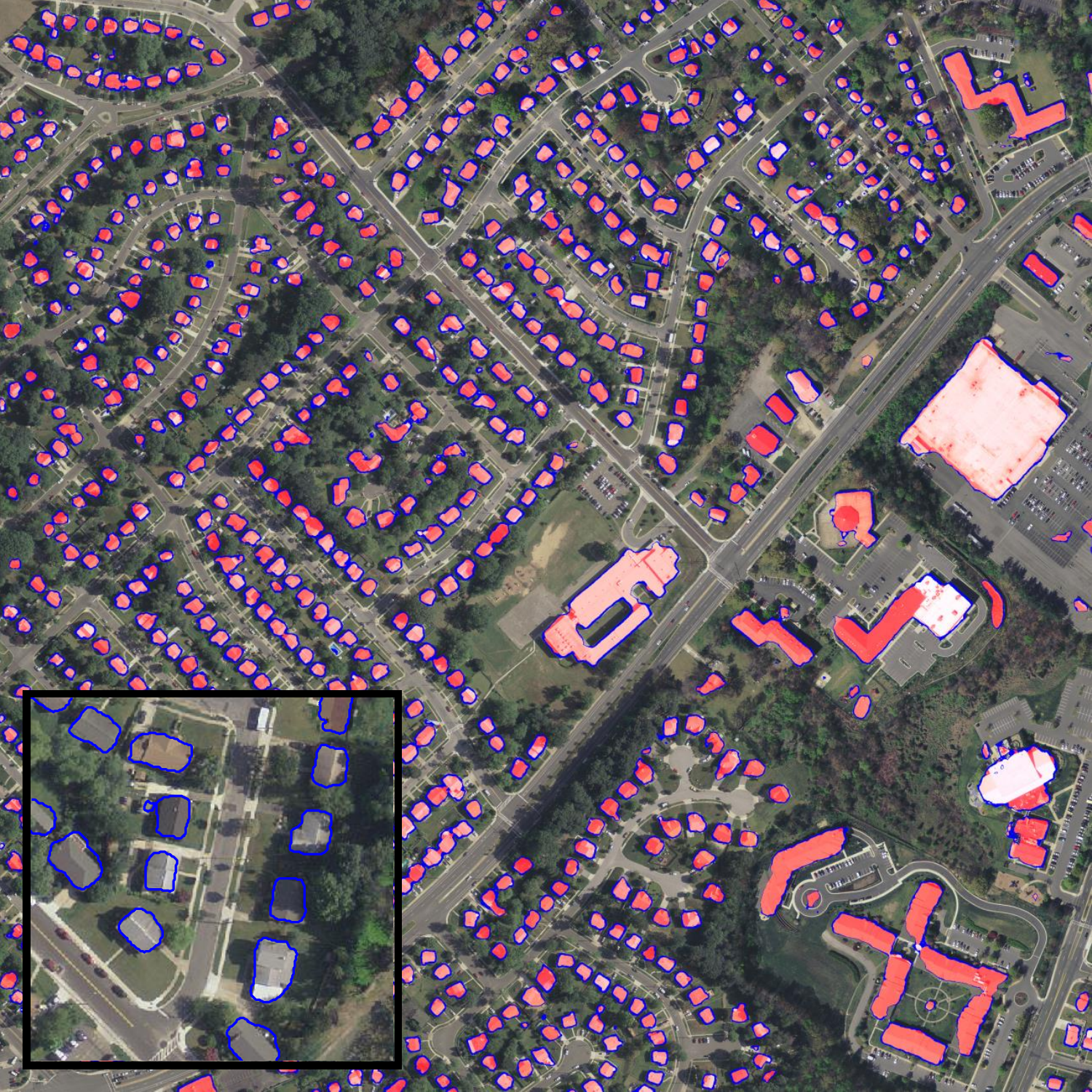}
\hspace{2mm}
\includegraphics[width=0.48\textwidth]{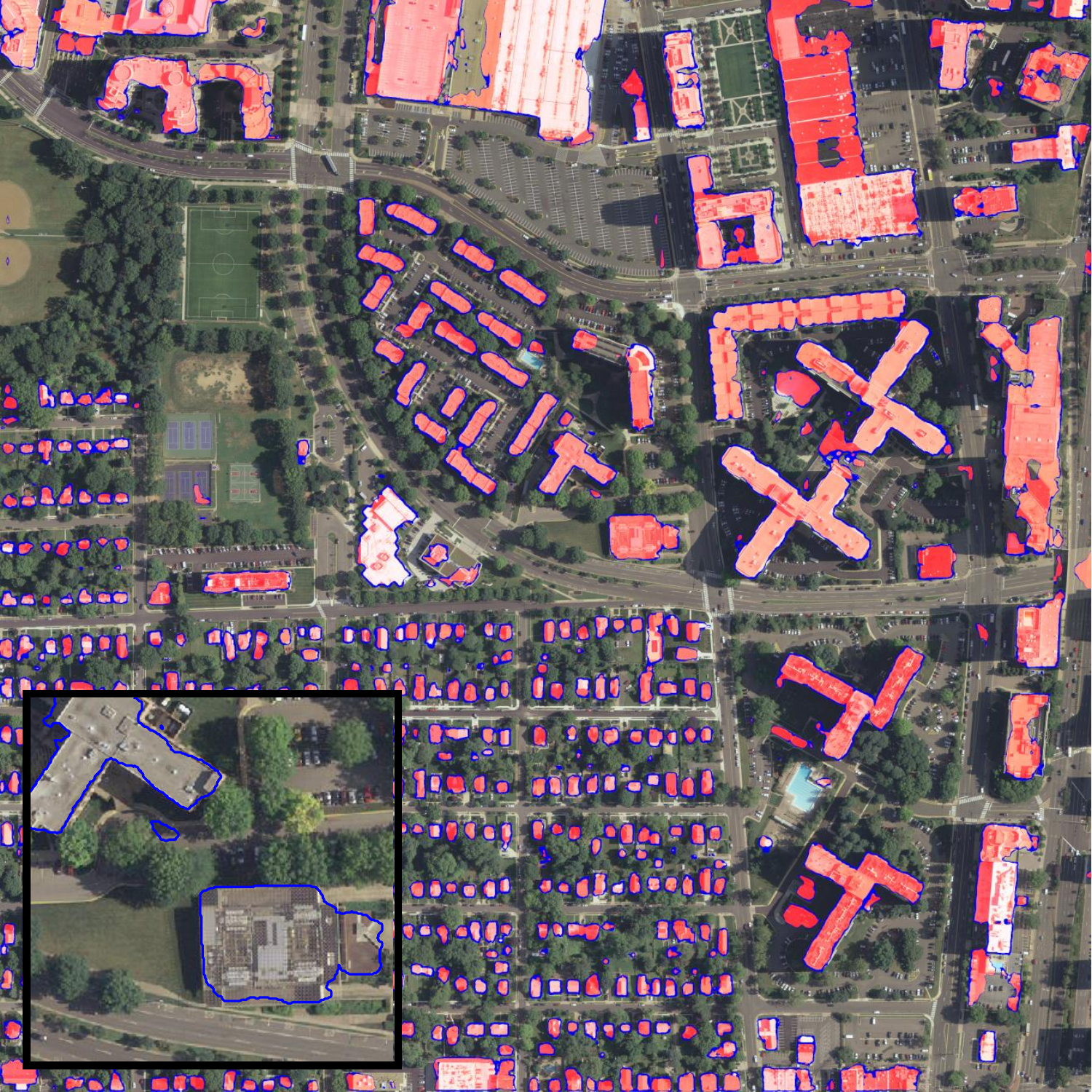}

\end{center}
\caption{Example results of two D.C. images. Transparent red regions indicate buildings extracted by our method, and blue pixels show detected building boundaries. For each image, a zoomed in view is provided with building boundaries marked. \label{fig:res1}}
\end{figure*}

Each of ten test images is input to the trained network without any pre-processing or tiling. Results of two images are shown in Fig.~\ref{fig:res1}, where pixels with output values between -0.5 and 0.5 are marked in blue and pixels with output values larger than 0.5 are marked in transparent red. To demonstrate the network capability, we present raw network output without any processing except the simple thresholding. As can be seen, images contain a mixture of various types of buildings in heterogeneous environments, which are very challenging for building extraction. The network identifies most of the buildings regardless of large variations. Zoomed in views are also provided in Fig.~\ref{fig:res1}, demonstrating boundary localization quality.

The precision and recall measures are commonly used to quantitatively evaluate building extraction results. Precision is the percentage of the true positive pixels among building pixels detected by the algorithm, and recall the percentage of the true positive pixels among building pixels in ground truth. The average precision and recall for test images are 0.81 and 0.80, respectively. 

\begin{figure*}
\begin{center}
\includegraphics[width=0.48\textwidth]{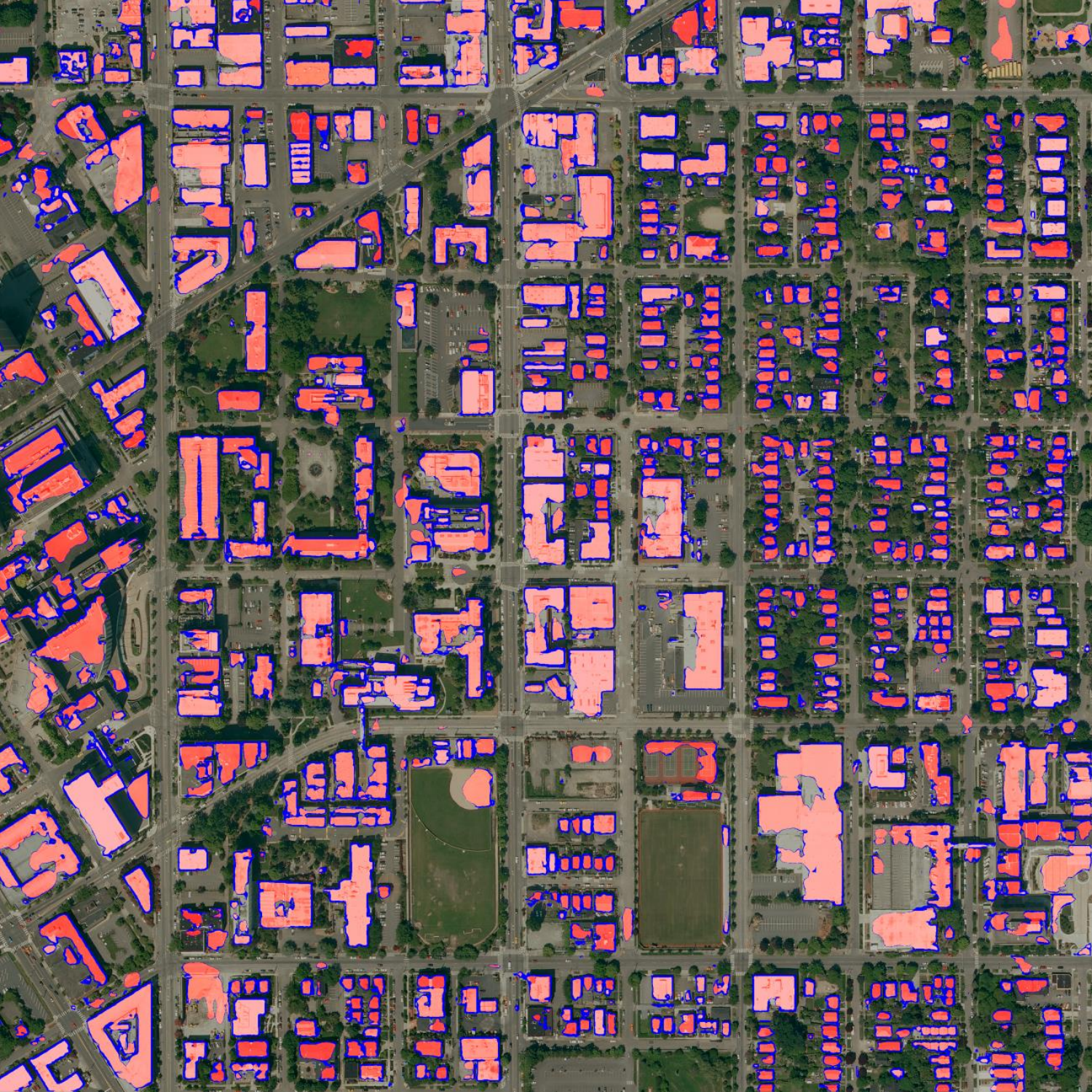}
\hspace{2mm}
\includegraphics[width=0.48\textwidth]{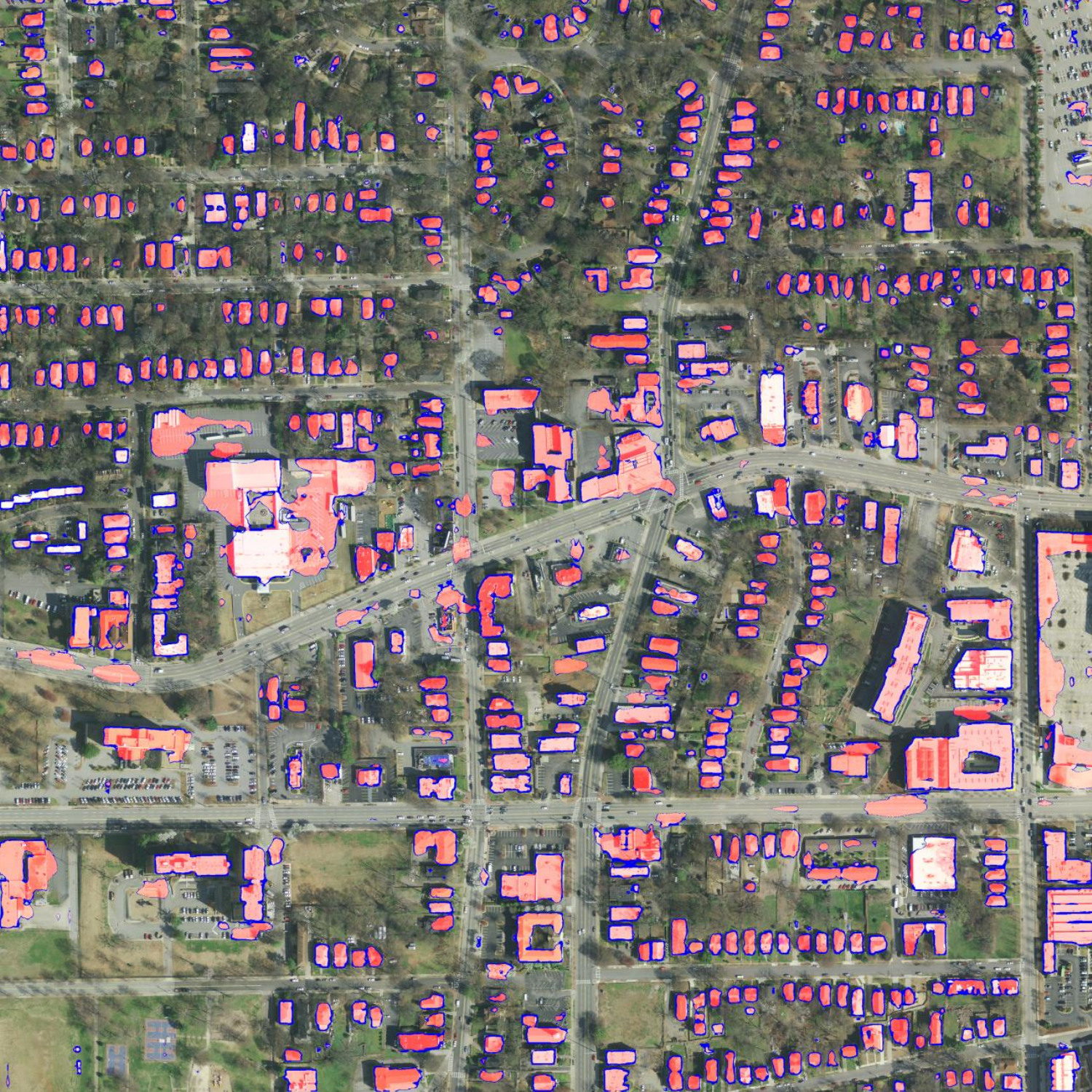}
\end{center}
\caption{Example results of images covering different cities. Left: Seattle, WA. Right: Atlanta, GA. \label{fig:res2}}
\end{figure*}

To further assess the generalization ability of the network, we create even more challenging test cases. We select five $3000 \times 3000$ images covering different U.S. cities, including Atlanta, Cincinnati, Dallas, Seattle, and San Francisco. The images are of 0.3 meter resolution but captured by different organizations at different time. We apply the trained network to the images and evaluate raw output. Fig.~\ref{fig:res2} illustrates results of two images from Seattle and Atlanta. Compared to the D.C. images, the images exhibit different spectral and spatial characteristics due to changes in atmospheric effects, imaging sensors, illumination, geographic features, etc. Two particularly noteworthy differences are 1) shadow directions in both images are different from that in the  D.C. images, and 2) in the Atlanta image a large number of buildings are severely occluded by vegetation, which are not observed in training data. The results show that the network generalizes well to those unseen conditions. We use as ground truth building footprints from OpenStreetMap with manual correction on inconsistent ones.  Despite large discrepancy between training images and test images, the network reaches an average precision of 0.74 and recall of 0.70 for five images.

\subsection{Comparison}
We cannot find any commercial software or software from published work that produces reasonable extraction results on our test data. We choose to conduct comparison on building detection. Detection is an easier problem than extraction because it aims to locate individual buildings without providing their spatial extents. Therefore, we can expect more reliably performance from existing methods. We select an influential method proposed by Sirmacek and Unsalan \cite{sirmacek2009}, which will be referred to as the SU method. This method constructs graphs based on detected keypoints. Based on provided templates, buildings are identified via subgraph matching. The code distributed by the authors is applied to the two images in Fig.~\ref{fig:res1} with templates selected from the images. We follow the evaluation strategy in \cite{sirmacek2009}, which regards a detected location inside buildings as true detection (TD) and otherwise false alarm (FA). To compare on an equal footing, we convert extraction output to detection output by using the mass center of each extracted building. Table 1 summarizes the performance of two methods. As can be seen, our method finds significantly more buildings than the SU method with a small number of false alarms. 

Since only simple local operations (convolution and maximization) are performed, our network is fast to evaluate, which takes around 1 minute to extract buildings on one test image. In contrast, the SU method takes more than 20 minutes to complete building detection. 

\begin{table}
\begin{center}
\caption{Comparison on building detection}
\begin{tabular}{c C{1cm} C{1cm}}
\hhline{===}
Method  & TD & FA  \\ \hline
\multicolumn{3}{c}{Image 1} \\
\multicolumn{3}{c}{(813 buildings)} \\[0.5ex] 
SU method  & 321 & 51 \\ 
Proposed & 708 & 45 \\ 
\multicolumn{3}{c}{Image 2 } \\
\multicolumn{3}{c}{(624 buildings)} \\[0.5ex]
SU method & 258 & 47 \\ 
Proposed & 574 & 31 \\
\hhline{===}
\end{tabular}
\end{center}
\label{tab:comp}
\end{table}

\section{Concluding remarks}
\label{sec:ccl}
In this paper we have presented a new building extraction method that leverages the ConvNet framework and rich GIS data. We design a network with novel components including a structure integrating multi-layer information for pixel-wise classification and a unique output representation. The components are easy to implement and enable the network to learn hierarchical features for segmenting individual objects. The trained system is tested on large real-world datasets and delivers accurate results with high efficiency. 

It is tempting to create a building extraction system that works at a global scale. Trained on data containing urban and suburban scenes in a particular city, the current system exhibits an outstanding generalization ability. However, we indeed observe performance degradation when applying the system to rural areas or different countries. To attain a universal building extractor, we need to address issues including how much labeled data are needed to handle worldwide geographic variations, what is the optimal network configuration, whether multiple networks for different geographic regions are better options than a single planet-scale network, etc. We are currently investigating these issues.   


The method proposed in this paper segments semantic objects (buildings) in complex scenes, which is a special case of figure-ground segmentation. Experimental results show that with sufficient labeled data our ConvNet model effectively separates foreground objects from their background in unseen data. We believe that this technique potentially provides a generic solution to figure-ground segmentation and benefits many challenging tasks across different domains. 
 
\appendices
\section{Calculation of receptive field size}
The receptive field of a unit in a convolutional network consists of units in the input layer that can pass activation to that unit. Note that local receptive fields refer to the connected units in the immediately preceding layer, while we are interested in connected units in the input layer. Knowing the receptive field size is important for analyzing network capacity. However, it is not straightforward to find out the receptive field size of a unit in a ConvNet with multiple stages. Here I provide a simple approach for this calculation. 

\begin{figure}
\begin{center}
\includegraphics[width=0.4\textwidth]{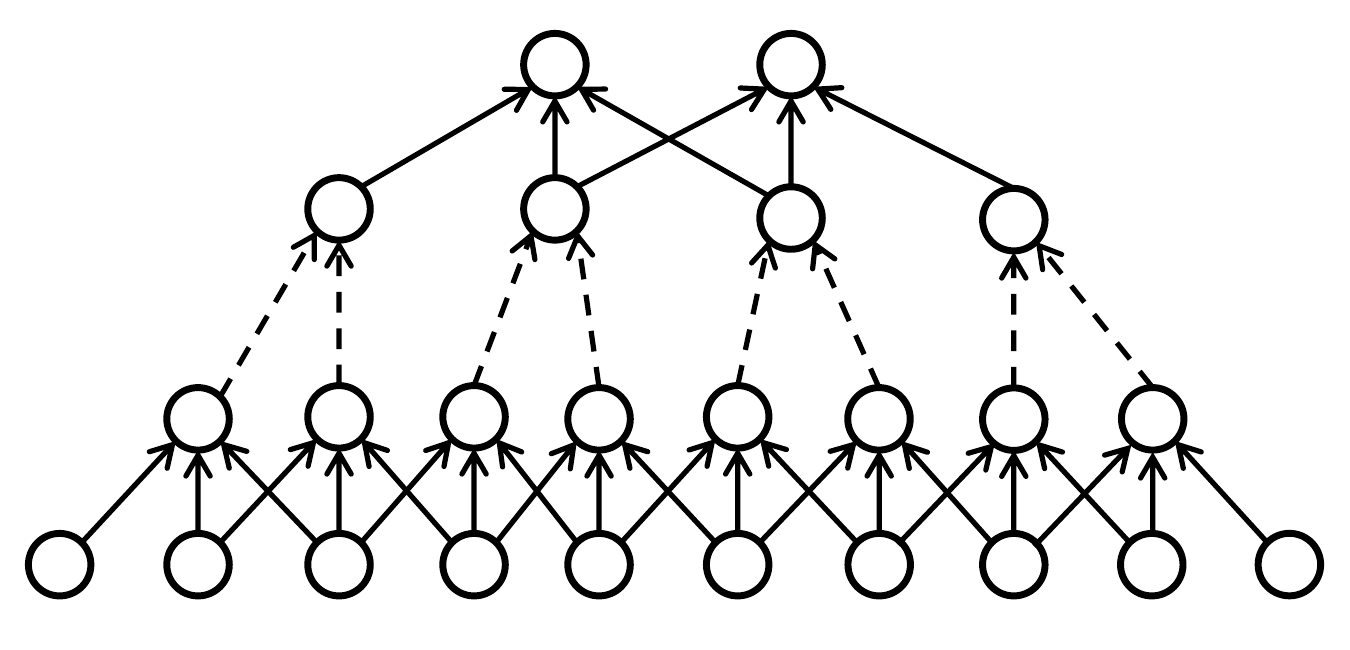}
\end{center}
\caption{An one-dimensional ConvNet. The input layer is followed by a convolutional layer with a filter of size 3, a pooling layer with a pooling region of 2, and a convolutional layer with a filter of size 3. Solid arrows represent convolution, and dashed arrows pooling. \label{fig:appdx1}}
\end{figure}

It is sufficient to analyze one-dimensional network, which can be conveniently extended to two-dimensional cases by conducting the same analysis for both dimensions. We assume that networks consist of convolutional layers and pooling layers, two main components in ConvNets. A small network is illustrated in Fig.~\ref{fig:appdx1}. To calculate the receptive field size of an output unit, an intuitive way is to start from the input layer and track connected units based on local receptive fields in each layer. However, since local receptive fields in a convolutional layer overlap with each other, with multiple layers it is difficult to count the connected units without drawing the entire network. 

There exist two rules regarding how unit numbers change between layers. 1) A convolutional layer decreases the unit number by $(s-1)$, where $s$ denotes the filter size. The decreased units are caused by border effects. 2) A pooling layer reduces the number of units by $p$ times, where $p$ is the size of pooling region. Two rules can be clearly observed in Fig.~\ref{fig:appdx1}. 

To find the receptive field size of a particular unit, we treat that unit and connected units in all previous layers as a sub-network and iteratively apply the two rules in a backward pass until obtaining the unit number in the input layer, i.e., the receptive field size. 
Formally, assuming an $m$-stage network with stage $i$ containing a convolutional layer with a filter of size $s_i$ and a pooling layer with a pooling region of $p_i$ ($p_i$ equals to 1 if there is no pooling layer), the relationship between unit numbers at stages $i$ and $i+1$ can be expressed by $R(i) = p_i R(i+1)  + (s_i -1) $. Knowing that the unit number at the last stage is $R(m)$ =1, the equation is applied iteratively until we have the unit number of input layer $R(0)$.

\bibliographystyle{ieee}
\bibliography{bldgdtc}

\end{document}